\documentclass{article}

\usepackage{arxiv}

\usepackage[utf8]{inputenc} % allow utf-8 input
\usepackage[T1]{fontenc}    % use 8-bit T1 fonts
\usepackage[hidelinks]{hyperref}       % hyperlinks
\usepackage{url}            % simple URL typesetting
\usepackage{booktabs}       % professional-quality tables
\usepackage{amsfonts}       % blackboard math symbols
\usepackage{nicefrac}       % compact symbols for 1/2, etc.
\usepackage{microtype}      % microtypography
\usepackage{cleveref}       % smart cross-referencing
\usepackage{lipsum}         % Can be removed after putting your text content
\usepackage{graphicx}
\usepackage{natbib}
\usepackage{doi}
\usepackage{tabularx}
\usepackage{subcaption}

\title{CACTUS: Chemistry Agent Connecting Tool-Usage to Science}

\newif\ifuniqueAffiliation
\ifuniqueAffiliation % Standard variant of author block
\author{ \href{https://orcid.org/0000-0000-0000-0000}{\includegraphics[scale=0.06]{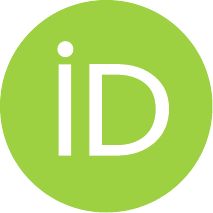}\hspace{1mm}David S.~Hippocampus}\thanks{Use footnote for providing further
		information about author (webpage, alternative
		address)---\emph{not} for acknowledging funding agencies.} \\
	Department of Computer Science\\
	Cranberry-Lemon University\\
	Pittsburgh, PA 15213 \\
	\texttt{hippo@cs.cranberry-lemon.edu} \\
	\And
	\href{https://orcid.org/0000-0000-0000-0000}{\includegraphics[scale=0.06]{orcid.pdf}\hspace{1mm}Elias D.~Striatum} \\
	Department of Electrical Engineering\\
	Mount-Sheikh University\\
	Santa Narimana, Levand \\
	\texttt{stariate@ee.mount-sheikh.edu} \\
}
\else
\usepackage{authblk}

\newbox{\orcid}\sbox{\orcid}{\includegraphics[scale=0.06]{orcid.pdf}} 
\author[1]{%
	\href{https://orcid.org/0000-0002-4146-7921}{\usebox{\orcid}\hspace{1mm}Andrew D.~McNaughton}%
 }
\author[1]{%
	\href{https://orcid.org/}{\usebox{\orcid}\hspace{1mm}Gautham Ramalaxmi}%
}
\author[1]{%
	\href{https://orcid.org/0000-0002-5571-7418}{\usebox{\orcid}\hspace{1mm}Agustin Kruel}%
}
\author[1]{%
	\href{https://orcid.org/0000-0002-1953-2272}{\usebox{\orcid}\hspace{1mm}Carter R. Knutson}%
}
\author[1]{%
	\href{https://orcid.org/0000-0002-2355-1484}{\usebox{\orcid}\hspace{1mm}Rohith A. Varikoti}%
}
\author[1]{%
	\href{https://orcid.org/0000-0001-6713-2129}{\usebox{\orcid}\hspace{1mm}Neeraj ~Kumar\thanks{\texttt{neeraj.kumar@pnnl.gov}}}%
}
\affil[1]{Pacific Northwest National Laboratory, Richland, WA 99354}
\fi

\renewcommand{\shorttitle}{CACTUS}

\hypersetup{
pdftitle={CACTUS: Chemistry Agent Connecting Tool-Usage to Science},
pdfsubject={},
pdfauthor={Andrew D.~McNaughton, Gautham Ramalaxmi, Agustin Kruel, Carter R. Knutson, Rohith A. Varikoti, Neeraj ~Kumar},
pdfkeywords={LLMs, Tool-Augmented Language Model, Molecular Discovery, Drug Design, Cheminformatics, Automated Molecular Design},
}

\begin{document}
\maketitle

%TC:break Abstract
%the command above serves to have a word count for the abstract
\begin{abstract}

Large language models (LLMs) have shown remarkable potential in various domains, but they often lack the ability to access and reason over domain-specific knowledge and tools. In this paper, we introduced CACTUS (Chemistry Agent Connecting Tool-Usage to Science), an LLM-based agent that integrates cheminformatics tools to enable advanced reasoning and problem-solving in chemistry and molecular discovery. We evaluate the performance of CACTUS using a diverse set of open-source LLMs, including Gemma-7b, Falcon-7b, MPT-7b, Llama2-7b, and Mistral-7b, on a benchmark of thousands of chemistry questions. Our results demonstrate that CACTUS significantly outperforms baseline LLMs, with the Gemma-7b and Mistral-7b models achieving the highest accuracy regardless of the prompting strategy used. Moreover, we explore the impact of domain-specific prompting and hardware configurations on model performance, highlighting the importance of prompt engineering and the potential for deploying smaller models on consumer-grade hardware without significant loss in accuracy. By combining the cognitive capabilities of open-source LLMs with domain-specific tools, CACTUS can assist researchers in tasks such as molecular property prediction, similarity searching, and drug-likeness assessment. Furthermore, CACTUS represents a significant milestone in the field of cheminformatics, offering an adaptable tool for researchers engaged in chemistry and molecular discovery. By integrating the strengths of open-source LLMs with domain-specific tools, CACTUS has the potential to accelerate scientific advancement and unlock new frontiers in the exploration of novel, effective, and safe therapeutic candidates, catalysts, and materials. Moreover, CACTUS's ability to integrate with automated experimentation platforms and make data-driven decisions in real time opens up new possibilities for autonomous discovery. The agent can design and prioritize experiments, analyze results, and iteratively refine its hypotheses, leading to more efficient and targeted exploration of chemical space.

%Moreover, CACTUS's ability to integrate with automated experimentation platforms and make data-driven decisions in real-time opens up new possibilities for autonomous discovery. The agent can design and prioritize experiments, analyze results, and iteratively refine its hypotheses, leading to more efficient and targeted exploration of chemical space.

%\lipsum[5]
\end {abstract}
%TC:break main
%the command above serves to have a word count for the abstract
% keywords can be removed
\keywords{LLMs \and Tool-Augmented Language Model \and Molecular Discovery \and Drug Design \and Cheminformatics \and Automated Molecular Design}

\section{Introduction} % Neeraj will assist, Agustin doing some additional lit review

Large Language Models (LLMs) are foundation models that are combined under a single paradigm to support various tasks or services. Despite being trained on vast corpora of data, these transformer-based LLMs have a limited understanding of the curated or parsed text.\citep{Chiesurin2023}. Current research has revealed the possibility of augmenting LLMs with tools that aid in efficiently solving various problems and tasks \citep{mialon2023,xu2023toolbench,qin2023toolllm}. Previous work has also shown that providing specific prompts, curated towards a specific task, can enhance the time and quality of the text generated by the models \citep{Cai2023}. Combining these two approaches is the Tool Augmented Language Model (TALM) framework, detailed in \citet{Parisi2022}, which outperforms existing models on the tasks it is configured for. However with any of these approaches, although the generated answers may appear correct, LLMs fail to reason or demonstrate subject knowledge as is typically demonstrated by humans \citep{huang2023large,kambhampati2024can}. Mistakes made by the model due to the statistical relationships it learned from data might appear in a similar way across different applications \citep{ontherisk}. If foundation models become integrated with important systems that leverage the foundation model’s ability to quickly adapt to many different tasks and situations, failures could result in significantly unwanted outcomes.

The resourceful LLMs like GPT4 \citep{gpt4}, LLaMA \citep{llama}, Gemma \citep{team2024gemma}, MPT \citep{MosaicML2023Introducing}, Falcon \citep{almazrouei2023falcon}, and Mistral \citep{jiang2023mistral} show improved performance over a range of activities \citep{chiang2024chatbot, zheng2023judging, hendrycks2020measuring}. Despite these strides, the inherent limitations of such models become apparent when faced with challenges that require access to dynamic, real-time, or confidential data, which remain inaccessible within their static training datasets. This gap underscores a critical need for LLMs to evolve beyond their current capacities, leveraging external APIs to fetch or interact with live data, thereby extending their utility in real-world applications \citep{Parisi2022}. In the domain-specific applications, particularly within the chemical, biological and material sciences, the limitations of LLMs are even more pronounced. The intricate nature of chemical data coupled with the dynamic landscape of drug discovery and development, presents a complex challenge that pure computational models alone cannot address effectively. Recognizing this, the integration of cheminformatics tools with the cognitive and analytical ability of LLMs offers a promising pathway. 

At the forefront of this transformation are Intelligent Agents, autonomous entities capable of designing, planning, and executing complex chemistry-related tasks with exceptional efficiency and precision \citep{agent1}. These systems are not only capable of utilizing a variety of LLMs for specific tasks but also adept at employing APIs and internet search tools to gather relevant material and data. For example, integrating an Agent into large, tool-based platforms such as KNIME \citep{KNIME} or Galaxy \citep{Galaxy} could form a natural language interface between the user and their analysis. By acting as intermediaries, these Agents could significantly streamline the process of scientific discovery and autonomous experimentation with or without human in the loop. Towards that end and taking inspiration from ChemCrow\citep{chemcrow}, an LLM-assisted chemistry synthesis planner, we have developed an \textit{ Intelligent Cheminformatics Agent} focused on assisting scientists with \textit{de novo} drug design and molecular discovery. Cheminformatics focuses on storing, retrieving, analyzing, and manipulating chemical data. It provides the framework and methodologies to connect computational linguistics with chemical science.  This synergistic approach aims to leverage the strengths of both domains by facilitating a more comprehensive and effective exploration of therapeutic compounds, streamlining the drug development process, and ultimately accelerating the discovery from conceptualization to clinical application. In this work, we developed CACTUS (Chemistry Agent Connecting Tool Usage to Science) an LLM-powered agent that possesses the ability to intelligently determine the most suitable tools for a given task and the optimal sequence in which they should be applied, effectively optimizing workflows for chemical research and development. 

The implications of these intelligent agents are far-reaching. They enable the autonomous operation of complex tasks from data analysis to experimental planning, hypothesis generation, testing, and push the boundaries of what can be achieved through computational chemistry. The synergistic relationship between human intelligence, artificial intelligence, and specialized software tools holds the potential to transform the landscape of drug discovery, catalysis, material science, and beyond. This relationship and combination of domains makes the molecular discovery process more efficient, accurate, and innovative. As we stand on the precipice of this new era in cheminformatics, the integration of LLMs and computational tools through intelligent agents like CACTUS promises to unlock a future where the limits of scientific discovery are bound only by the depths of our imagination.

\section{Methods} % Andrew will focus on this

Tool-augmented language models consist of two major components: external tools and language models. This section will discuss the approaches used to implement the language model agent and provide a focused look at the tools used. We will also go into great detail about the strategies used when prompting our agent and how we performed benchmarking. Each of these steps is a critical component of forming a complete intelligent agent able to solve a wide range of problems with the added ability of quick model swapping.

\subsection{The Agent}
%General Methods (discuss langchain, LLMs, tool construction)
An important consideration when building a TALM is the framework in which it will be implemented. We have selected the commonly used open-source platform, LangChain \citep{Chase_LangChain_2022}, for this purpose. This framework simplifies the integration of prompts with LLMs through a comprehensive set of pre-built Python modules known as "chains". It also provides convenient integration with popular LLM hosting/inference platforms such as the OpenAI API and HuggingFace Transformers \citep{Wolf_HF}. CACTUS utilizes LangChain's implementation of a custom MRKL agent \citep{karpas2022mrkl} which can be broken into 3 parts: tools, LLMChain, and agent class. The tools in this instance are a collection of cheminformatics helper functions that wrap well-known Python libraries into well-described tools for an agent to use. These tools are explained in much more detail in Section \ref{sec:Tools}. The LLMChain is a LangChain specific feature that helps chain the tools and the agent together. This is the prompt provided to the LLM when running any inference and helps to instantiate the model and parse the user input. In CACTUS, we provide a prompt that guides the agent to answer cheminformatics questions by describing the typical steps involved in answering such questions. The last requirement for CACTUS is the agent class. These are also LangChain implemented functions that are used to interpret the user input after the initial prompt and make decisions on which actions to take to best solve the question. CACTUS sticks with a general purpose implementation of the zero-shot agent class that uses the ReAct \citep{Yao2022} framework to determine which tool to use from the tool's description. This combination of tools, LLMChain, and zero-shot agent makes CACTUS an extensible LLM tool that can quickly integrate new tools to solve a range of cheminformatics questions.

Here, we introduce mathematical formulation to describe the key components and processes of the CACTUS framework:

Let's consider $T = {t_1, t_2, \ldots, t_n}$ the set of cheminformatics tools available to CACTUS as discussed above, where each tool $t_i$ is a function that takes an input $x_i$ and produces an output $y_i$:
\begin{equation}
t_i(x_i) = y_i
\end{equation}
The LLMChain is represented as a function $L$ that takes a user input $u$ and a set of tools $T$ as input, and outputs a sequence of actions $A = {a_1, a_2, \ldots, a_m}$:
\begin{equation}
L(u, T) = A
\end{equation}
Each action $a_i$ in the sequence $A$ corresponds to the application of a specific tool $t_j$ on an input $x_j$, resulting in an output $y_j$:
\begin{equation}
a_i = t_j(x_j) = y_j
\end{equation}
The zero-shot agent class is  modeled as a function $Z$ that takes the user input $u$, the set of tools $T$, and the LLMChain output $A$ as input, and produces a final output $o$:
\begin{equation}
Z(u, T, A) = o
\end{equation}
The final output $o$ is the result of executing the sequence of actions $A$ determined by the LLMChain, given the user input $u$ and the available tools $T$.
Here, The ReAct framework used by the zero-shot agent class was  represented as a function $R$ that takes the user input $u$, the set of tools $T$, and the tool descriptions $D = {d_1, d_2, \ldots, d_n}$ as input, and outputs the most appropriate tool $t_k$ to use:
\begin{equation}
R(u, T, D) = t_k
\end{equation}

This combination of cheminformatics tools, LLMChain, and zero-shot agent makes CACTUS an extensible LLM tool that can quickly integrate new tools to solve a range of cheminformatics questions.

\begin{figure}[ht]
    \centering
    \includegraphics[width=0.75\textwidth]{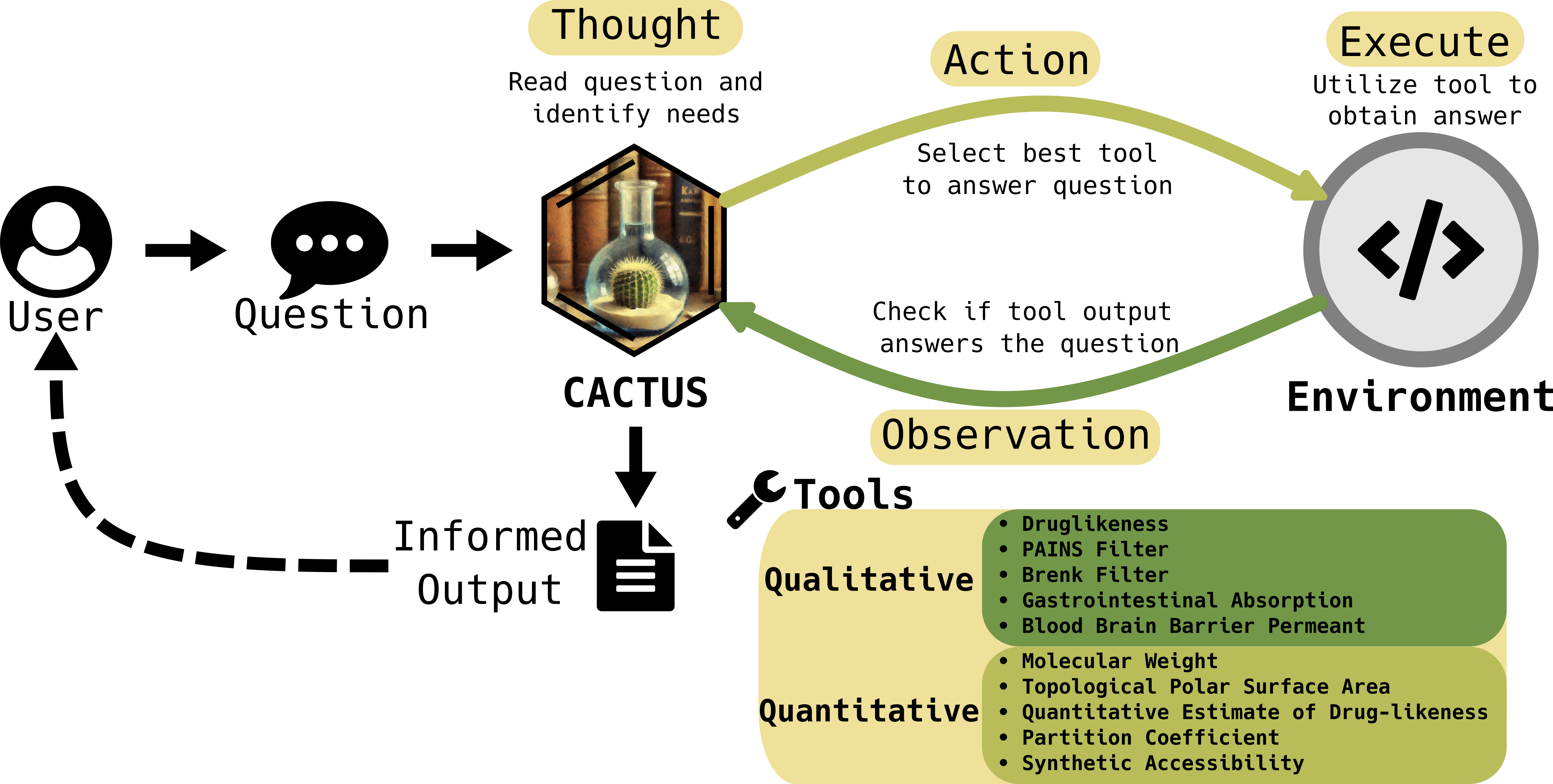}
    \caption{General workflow of the CACTUS Agent that details how the LLM interprets an input to arrive at the correct tool to use to obtain an answer. Starting from the user input, CACTUS follows a standard "Chain-of-thought" reasoning method with a Planning, Action, Execution, and Observation phase to obtain an informed output }
    \label{fig:workflow}
\end{figure}

%\subsubsection{Prompting Strategy}

\subsection{Cheminformatics Tools} % Rohith & Andrew will work on this
\label{sec:Tools}

For the purpose of creating a robust LLM agent able to answer a variety of cheminformatics questions, CACTUS includes a wide range of tools integrating common functions found in Python libraries such as RDKit \citep{landrum2013rdkit} and SciPy \citep{virtanen2020scipy}, along with interfaces to databases such as PubChem \citep{kim2023pubchem}, ChEMBL \citep{davies2015chembl}, and ZINC \citep{irwin2020zinc20}. These tools allow for a chat-based analysis of molecules starting with a SMILES string and ending with information such as molecular descriptors, similarity, or absorption, distribution, metabolism, and excretion (ADME) attributes. The model consists of ten different tools providing information on various descriptors for any given chemical compound used as input. Table \ref{table:tool-table} contains a list of currently available tools that can assist in obtaining different physio-chemical properties and molecular descriptors of the input chemical compounds. This includes molecular weight, log of the partition coefficient (LogP), topological polar surface area (TPSA), quantitative estimate of drug-likeness (QED), and synthetic accessibility (SA) of the input chemical compounds. Moreover, using the BOILED-Egg method, CACTUS can also estimate the pharmacokinetic properties like blood-brain barrier permeability and gastrointestinal absorption of any given chemical compound \citep{daina2016boiled}. Our model also implements drug-likeness, PAINS, and Brenk filters to identify structural and toxicity alerts. All these tools in our model will assist in identifying and screening both currently available and new lead compounds. Currently restricted to using a simple SMILES as input, future releases will allow for varied user input (compound name, molecular formula, InChI key, CAS number, SMILES, ChEMBL ID, or ZINC ID) where the agent will first convert it to SMILES notation, and then used as input for the available tools.

\begin{table*}[ht]
    \centering
    \begin{tabularx}{0.65\textwidth}{l X}
        \toprule
        \textbf{Tool} & \textbf{Description} \\
        \midrule
        MolWt & Float $[0,\infty]$ - Molecular weight \\
        LogP & Float $[-\infty,\infty]$ - Predicted partition coefficient \\
        TPSA & Float $[0,\infty]$ - Topological Polar Surface Area \\
        QED & Float $[0,1]$ - Quantitative Estimate of Druglikeness \\
        SA & Float $[1,10]$ - Synthetic Accessibility \\
        BBB Permeant & String $[Yes, No]$ - Is in "yolk" of BOILED-Egg model \\
        GI Absorption & String $[Low,High]$ - Is in "white" of BOILED-Egg model \\
        Druglikeness & Boolean - Passes Lipinski Rule of 5 \\
        Brenk Filter & Boolean - Passes Brenk filter \\
        PAINS Filter & Boolean - Passes PAINS filter \\
        \bottomrule
    \end{tabularx}
    \caption{Cheminformatics tools currently supported by CACTUS. These tools provide a comprehensive assessment of a molecular and  physicochemical properties. Apart from conversions between different molecular representations, all tools require input in the SMILES format. By leveraging these tools, CACTUS enables researchers to make informed decisions in the molecular discovery process and prioritize compounds with the most promising characteristics.}
    \label{table:tool-table}
\end{table*}

\subsection{Prompting Strategy}
One important aspect investigated was the significance of the prompt for the agent. Through the LangChain implementation of LLM agents, there is a default prompt that provides a generic instruction of what tools are available and what the task of the LLM is. However, this is not necessarily primed for understanding domain-specific information. To test the hypothesis we ran 2 scenarios: one where we left the default prompt unchanged and only included tool descriptions (Minimal Prompt), and one where we modified the prompt to align the agent more with the domain of chemistry (Domain Prompt). The belief is that a domain aligned prompt will steer the LLM towards better interpretation of the questions being asked, and therefore be more effective in answering user queries. Since we were using a wide range of LLMs for testing, the minimal prompt also included model-specific tokens so that we weren't unfairly evaluating models against the domain prompt.

\subsection{Benchmarking}
\label{sec:Benchmarking}
Evaluation of domain-specific TALMs can be a difficult task but we can follow the examples set by general benchmarking suites \citep{li2023,farn2023,Gentopia,xu2023toolbench}. Therefore, we rely on sets of questions that replicate the typical questions the agent would see and score how many the agent is able to answer correctly without requiring extra prompting effort from the user (i.e. having to rephrase the typed question to get a correct answer). To evaluate CACTUS we created sets of cheminformatics questions that test 3 sets of questions depending on the output of the tool. The first set is of \textit{qualitative} questions, and is represented by questions that return answers like Yes/No, or True/False. The second is \textit{quantitative}, which represents tools that return numerical values to be interpreted by the agent. The third is a combination of both qualitative and quantitative which we call \textit{full} or \textit{combined} set. Table \ref{table:benchmark} highlights examples of questions passed as user-input to the CACTUS agent. The qualitative and quantitative datasets each contain 500 questions, and the combined dataset contains 1000. Most tests will be done on the combined dataset as we want to test the LLM agent's ability to perform a diverse set of tasks.

\begin{table*}[ht]
    \centering
    \begin{tabularx}{0.85\textwidth}{X X l}
    \toprule
    \multicolumn{3}{c}{Qualitative Questions} \\ \cmidrule(lr){1-3}
    Question   & Step   & Answer \\ \midrule
    Does CCON=O pass the blood brain barrier? & Use BBB Tool w/ SMILES  & Yes   \\
    What is the GI absorption of C\#C? & Use GI tool w/ SMILES  & Low                \\ \midrule
    \multicolumn{3}{c}{Quantitative Questions} \\ \cmidrule(lr){1-3}
    Question & Step & Answer \\ \midrule
    What is the QED of CCCC=O?   & Use QED Tool w/ SMILES   & 0.44    \\
    What is the TPSA of C(CS)O  & Use TPSA Tool w/ SMILES & 20.23 \\ \bottomrule
    \end{tabularx}
    \caption{Table demonstrating examples of the questions asked of the CACTUS agent in the cheminformatics benchmark used in this paper.}
    \label{table:benchmark}
\end{table*}

\section{Results and Discussion}
The implementation of CACTUS represents a significant step forward in the field of cheminformatics, offering a powerful and flexible tool for researchers and chemists engaged in molecular discovery and drug design. The benchmarking studies conducted on various 7b parameter models demonstrate the robustness and efficiency of the CACTUS framework, highlighting its potential to streamline and accelerate the drug discovery process as an example.

\subsection{Benchmarking and Performance Evaluation
}

\begin{figure}[h]
  \centering
  \begin{subfigure}[b]{0.8\textwidth}
    \centering
    \includegraphics[width=\textwidth]{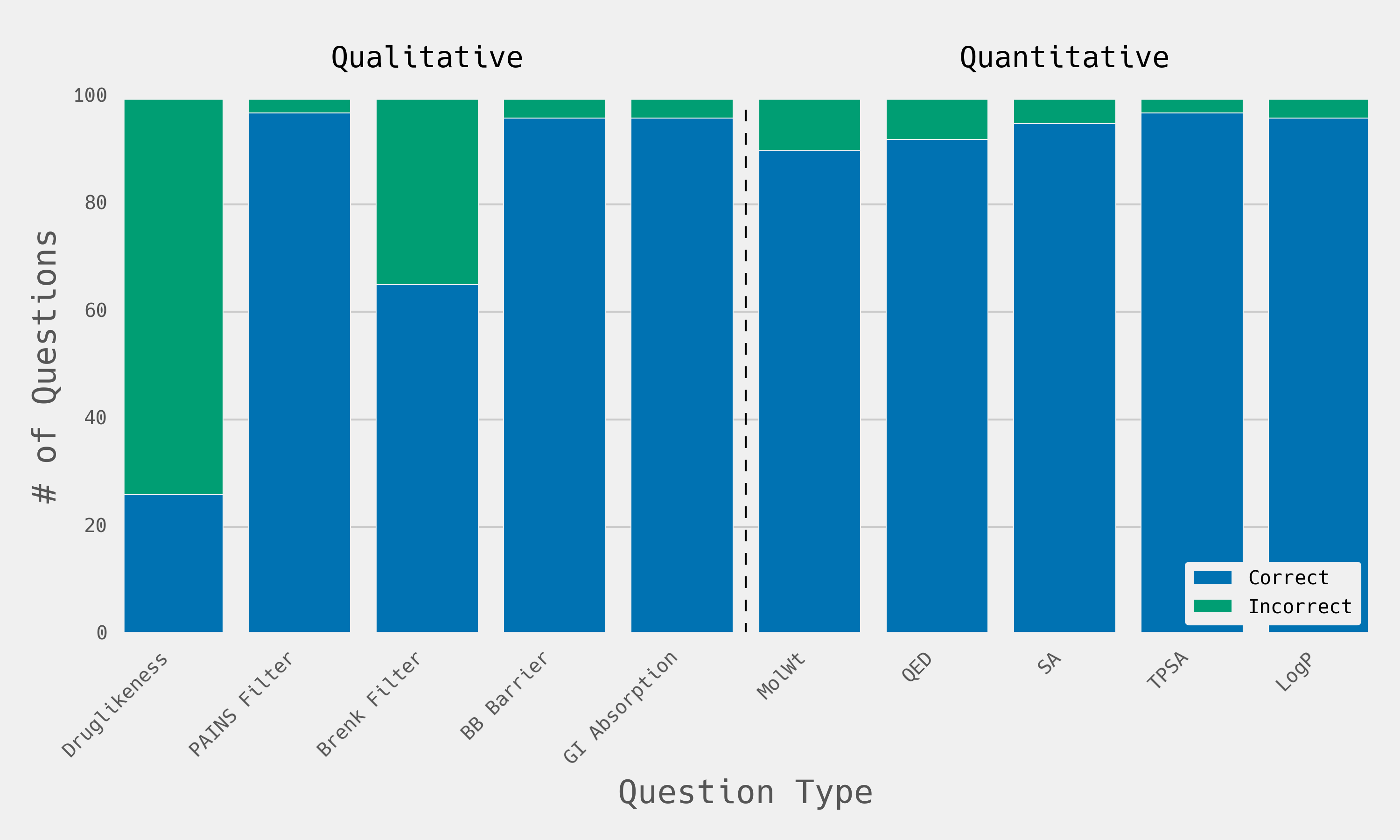}
    \caption{Benchmark performance on the Gemma-7b model with a minimal prompt on each of the 10 question types.}
    \label{fig:no_prompt}
  \end{subfigure}
  \hfill
  \begin{subfigure}[b]{0.8\textwidth}
    \centering
    \includegraphics[width=\textwidth]{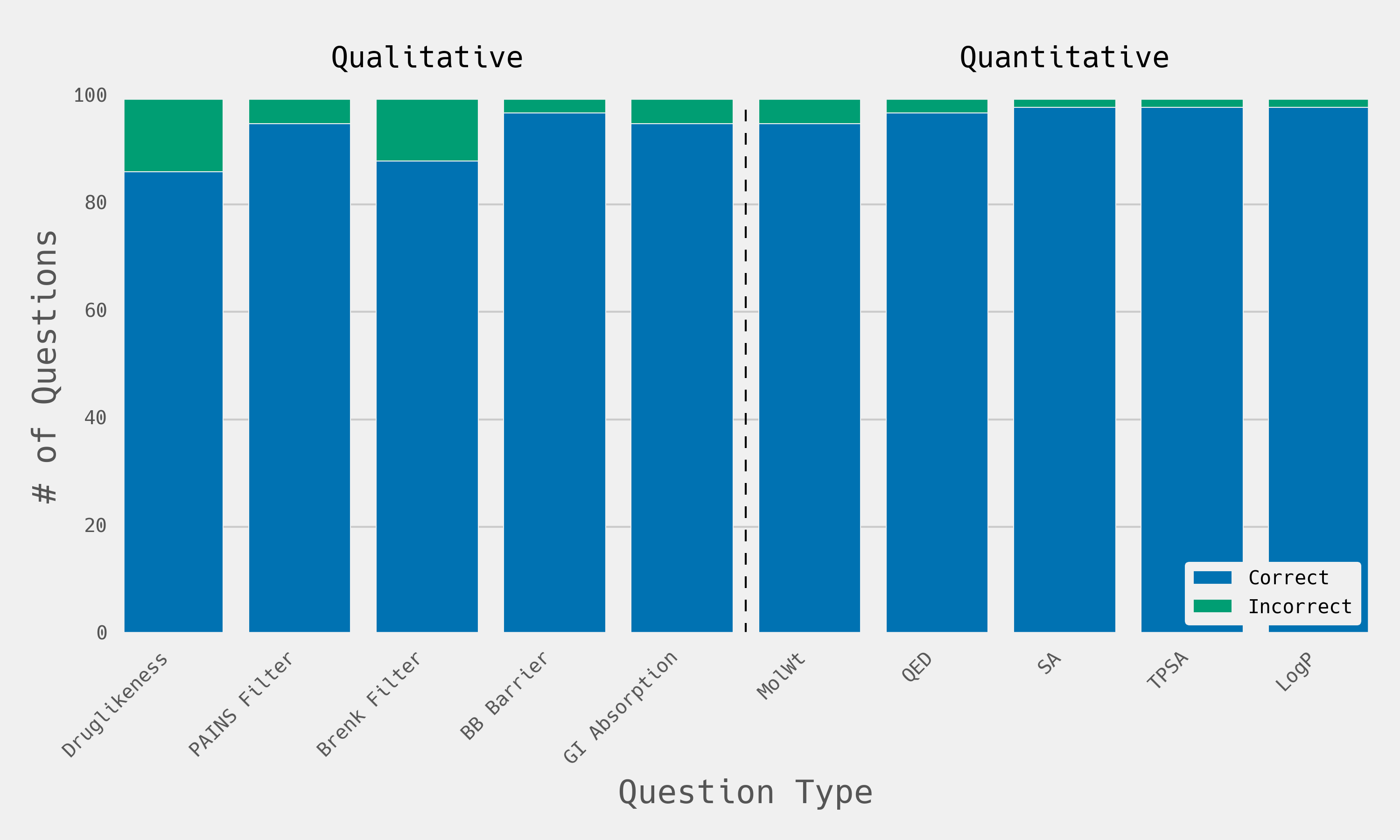}
    \caption{Benchmark performance on the Gemma-7b model with a domain prompt on each of the 10 question types.}
    \label{fig:prompt}
  \end{subfigure}
  \caption{Comparison of the Gemma -7b model with different prompting strategies on the full question set benchmark shows significant improvement in the qualitative question set when comparing the minimal prompt (Figure \ref{fig:no_prompt}) to the domain prompt (Figure \ref{fig:prompt}), while demonstrating similar performance in the quantitative question set.}
  \label{fig:overall}
\end{figure}

The performance of CACTUS was evaluated using a comprehensive set of 1000 questions, covering 10 different tools (Table \ref{table:tool-table}, with and without the domain prompt on each 7b parameter model as shown in the Figure \ref{fig:overall}. Correct answers were scored as \textit{correct}, while wrong answers, inability to converge on an answer, or inability to use the provided tool correctly were marked as \textit{incorrect}. In this paper, we did not differentiate between incorrect tool usage and simply providing a wrong answer. Any answers that did not coherently address the question were considered incorrect. We accepted correct answers that contained additional formatted text after the correct answer, although this is not the preferred format. This additional information can be programmatically removed before returning the response to the user, or further prompts can be engineered to reduce additional text. Each type of question in the full question set was asked 100 times, resulting in 10 types of questions corresponding to the 10 tools provided in Table \ref{table:tool-table}. This approach allowed us to identify which tools posed a greater challenge for the model, and where improvements to either the tool description or model prompt could be made. 

The results shown in Figure \ref{fig:overall} highlight the importance of domain-specific prompting in improving the accuracy of the model's responses; particularly for qualitative questions. This finding aligns with recent research emphasizing the role of prompt engineering in enhancing the performance of language models \citep{liu2023gpt}.

\begin{figure}[!ht]
    \centering
    \includegraphics[width=1\linewidth]{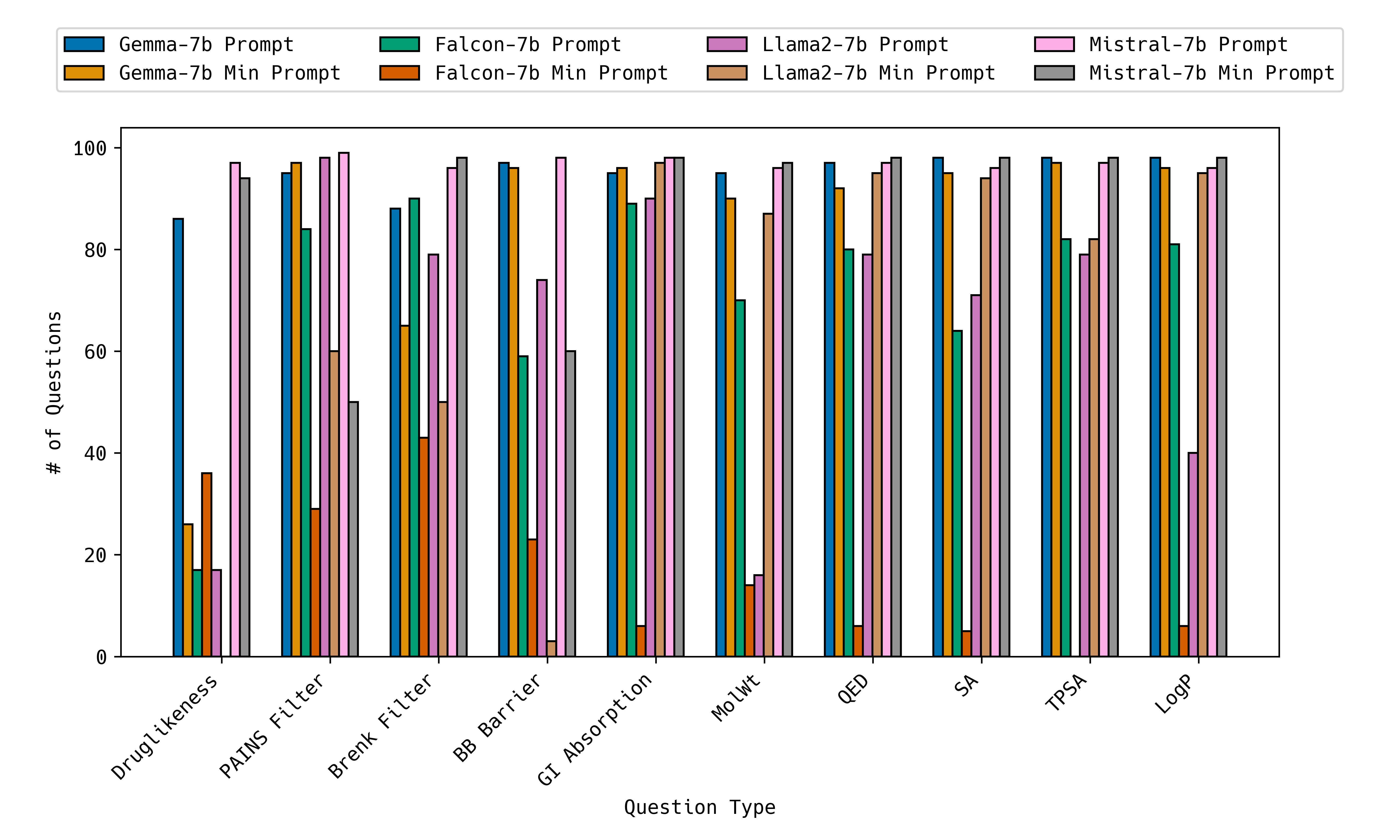}
    \caption{Comparison of model performance among 7B parameter models using minimal and domain-specific prompts. The Gemma-7b and Mistral-7b models demonstrate strong performance and adaptability across prompting strategies, highlighting their potential for widespread applicability in various computational settings, from high-performance clusters to more modest research setups.}
    \label{fig:model_comparison}
\end{figure}

In the progression of AI and its applications in scientific inquiry, it is crucial to analyze the comparative effectiveness of various models in handling domain-specific tasks. The benchmarking analysis presented in Figure \ref{fig:model_comparison} offers significant insights into the performance of different language models when prompted with both minimal and domain-specific information. A comprehensive review of the performance data across the full spectrum of question types reveals that Gemma-7b and Mistral-7b models showcase robustness and versatility, performing admirably regardless of the nature of the prompt. Their consistent accuracy across different types of questions ranging from physiochemical properties like druglikeness and blood-brain barrier permeability to more complex metrics like quantitative estimate of drug-likeness (QED) highlight their reliability for a broad range of inquiries within the domain of molecular science. In contrast, models like Falcon-7b exhibit a noticeable disparity between performances with minimal and domain prompts. This variability suggests that Falcon-7b, while capable, may require more fine-tuned prompting to leverage its full potential effectively. The substantial difference in performance based on the prompt type points to an intrinsic model sensitivity to input structure and content, which can be pivotal in crafting effective inquiry strategies. Furthermore, the successful deployment of smaller models, such as Phi2 and OLMo-1b, on consumer-grade hardware (Figure \ref{fig:acc_time}) highlights the potential for democratizing access to powerful cheminformatics tools, enabling researchers with limited computational resources to harness the capabilities of CACTUS.

\subsubsection{Open Source Models in Varied Settings} 
This comprehensive model comparison and analysis has broader implications for the employment of open-source models in scientific environments. The ability of models to perform well with domain-specific prompts is particularly encouraging, as it implies that with proper configuration, open-source models can be highly effective tools. The adaptability demonstrated by the Gemma-7b and Mistral-7b models indicates their potential for widespread applicability across various computational settings, from high-performance clusters to more modest research setups. Moreover, the ability to effectively prompt open-source models opens the door to their use in a variety of scientific contexts. It allows researchers to customize models to their specific domain, potentially bridging the gap between generalized AI capabilities and specialized knowledge areas.

The flexibility and performance of these models have significant implications for scientific research, particularly in fields like synthetic organic chemistry and drug discovery. For researchers in these domains, the ability to utilize open-source models effectively can accelerate the discovery process, enhance predictive accuracy, and optimize computational resources. The insights from this benchmarking study provide a roadmap for selecting and tailoring models to specific research needs, thereby maximizing their utility in advancing scientific goals. The benchmarking study of the selected 7b parameter models serves as a testament to the progress in AI-driven research tools. It highlights the necessity of prompt optimization and the promise of open-source models in diverse scientific inquiries. The analysis underscores the potential of these models to become integral components in the computational chemist's toolkit, paving the way for innovative breakthroughs in molecular design and drug discovery.

\subsection{Hardware Performance and Model Efficacy}

\begin{figure}[h]
    \centering
    \includegraphics[width=1\linewidth]{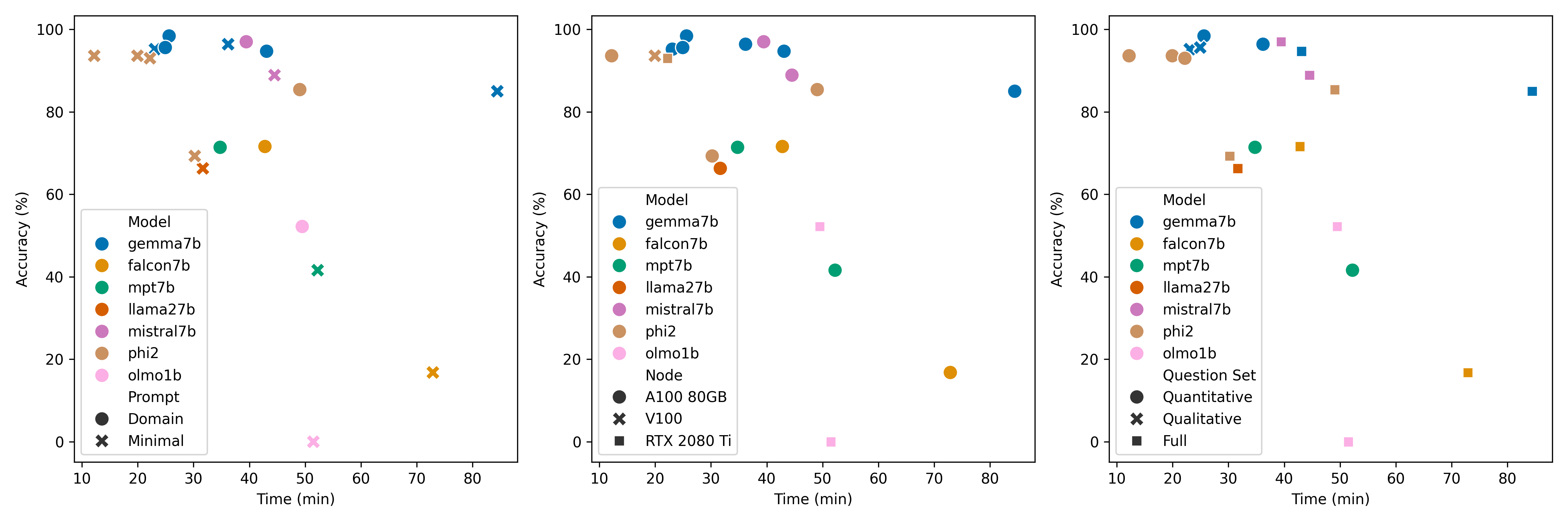}
    \caption{Comparison of model performance using accuracy and execution time as key metrics. The study evaluates various open-source models available on the HuggingFace including Gemma-7b, Falcon-7b, MPT-7b, Llama2-7b, and Mistral-7b, phi2 and olmo1b. Different combinations of conditions, such as model type (Vicuna, LLaMa, MPT), prompting strategy (minimal or domain-specific), GPU hardware (A100, V100, or consumer-grade), and benchmark size (small or large) were used to assess the model's capabilities.}
    \label{fig:acc_time}
\end{figure}

The deployment of CACTUS models through vLLM offers a significant advantage by optimizing performance across a variety of GPUs used for LLM inference. In our benchmarking studies we utilized three types of NVIDIA GPUs: the A100 80GB, V100, and RTX 2080 Ti. Our objective was to evaluate the performance of models under different combinations of model size, GPU type, and prompting strategy (minimal or domain-specific). The performance metric was determined by the inference speed in relation to the model's accuracy. Figure \ref{fig:acc_time} shows the summary of LLMs deployed under different conditions (GPU hardware used, prompt, and benchmark set used) and how well they performed. The efficiency of these models across diverse hardware highlights their potential for widespread implementation in a range of research settings. 

The models evaluated include Gemma-7b, Falcon-7b, MPT-7b, Llama2-7b, Mistral-7b, as well as two smaller models, Phi2 and OLMo-1b. The inclusion of these smaller models highlights the potential for successfully deploying models on local resources with limited computational power (e.g., consumer-grade GPUs like the RTX 2080 Ti) while still achieving accurate results. Overall, the model performance was found to be relatively quick on both the 500-question sets (Qualitative/Quantitative) and the 1000-question combined set (Full). One notable outlier was the Llama2-7b model with domain prompting, which took 185 minutes to complete the inference on the full dataset; however, its accuracy was similar to the minimally prompted version. This model is considered to be an outlier and therefor not included in Figure \ref{fig:acc_time}. A full list of the data used to plot these summary figures can be found in the Appendix. 

The most interesting outcome is that smaller models deployed on consumer grade hardware do not perform drastically worse than their larger parameter model counterparts. Looking at the performance of the Phi2 model (2.7B parameters), it quickly and accurately tackles the 500 question quantitative benchmark with similar performance regardless of the GPU used with the A100 80GB version unsurprisingly as the fastest. Another interesting outcome is the performance of the OLMo-1b parameter model on the combined question set and the RTX 2080 Ti GPU. While unable to obtain any correct answers for the minimal prompt, it jumps up to a surprising 52.2\% accuracy when provided a domain prompt. These results are promising that these smaller models can be deployed locally by users and still be able to interpret questions, possibly by providing more specialized prompts.

In general, inference time increased as question set size increased (e.g., from quantitative/qualitative to full) , while accuracy tended to decrease with longer inference times. Domain prompts achieved faster inference and accuracy than minimal prompts for models like Falcon-7b, MPT-7b, and Mistral-7b. However, there was an exception in the case of the Phi2 model on the full question set, where the minimal prompt resulted in faster inference but lower accuracy.

The hardware performance analysis highlights the importance of considering the interplay between model size, GPU capabilities, and prompting strategies when deploying CACTUS models for molecular property prediction and drug discovery. The ability to achieve accurate results with smaller models on consumer-grade hardware opens up the possibility of wider adoption and accessibility of CACTUS for researchers with limited computational resources. Furthermore, the impact of domain-specific prompting on both inference speed and accuracy emphasizes the need for carefully designed prompts tailored to the specific application domain.
As CACTUS continues to evolve and integrate with other computational tools and autonomous discovery platforms, optimizing hardware performance will remain a critical consideration. Future research should explore the development of more efficient algorithms and architectures (energy efficiency) for deploying CACTUS models on a variety of hardware configurations, ensuring that the benefits of this powerful tool can be realized across a wide range of research settings and computational resources.

\subsection{Issues Encountered and Resolutions}

During the development and benchmarking of CACTUS agent using open-source models and the LangChain framework, several key challenges were identified. These issues, along with the solutions implemented, provide valuable insights for researchers and developers working on similar workflows.

One of the primary issues encountered was the slow inference speed when hosting open-source language models locally on machines utilizing CPUs. Most APIs quickly provide inference results when making calls and this is not something locally hosted models typically replicate well, especially when running on CPUs over GPUs. For this work, we initially used models from HuggingFace and deployed through the HuggingFace Pipelines python package. This allowed us to serve models, but the inference time was quite slow when wrapped in the LangChain agent. To address this, we began utilizing vLLM to host HuggingFace models instead. This substantially decreased our inference time, and allowed for API-like response times from models, even those hosted on less powerful consumer grade GPU hardware.

The second major challenge was related to prompt engineering. Our results shown previously highlight that for some models the prompt has a great effect on not only the model accuracy, but the inference time. We spent a good amount of time trying to hone our prompting strategy to yield consistently accurate and efficient results with mixed effect. We ended up needing specialized prompts for each open-source LLM we used, as some were fine-tuned much differently than others and required a very specific prompt style to return usable results.

These challenges highlight the need for continued research and development in the areas of model deployment and prompt engineering. Future work will be focused on optimizing the deployment of open-source models on various hardware configurations, including CPUs and GPUs, to ensure that CACTUS can be efficiently utilized across a wide range of computational resources. This may involve the development of novel algorithms and architectures that can better leverage the capabilities of different hardware setups, as well as the creation of more user-friendly tools and frameworks for model deployment and management. In terms of prompt engineering, the development of standardized prompt templates and best practices for prompt engineering in the context of molecular property prediction and drug discovery could help streamline the development process and improve the consistency of results across different models and datasets.

\subsection{Future Outlook - Molecular Design}

CACTUS has already demonstrated its potential in estimating basic metrics for input chemical compounds, but its future lies in its evolution into a comprehensive, open-source tool specifically designed for chemists and researchers working on therapeutic drug design and discovery. This will be achieved by the integration of physics-based molecular AI/ML models, such as 3D-scaffold \citep{joshi20213d}, reinforcement learning \citep{McNaughton2022}, and graph neural networks (GNNs) \citep{knutson2022decoding} accompanied with molecular dynamics simulations, quantum chemistry calculations, and high-throughput virtual screening \citep{joshi2021quantum, knutson2022decoding, joshi2023ai, varikoti2023integrated, Joshi2021}. Such capabilities are essential for accurately modeling molecular interactions and predicting the efficacy and safety of potential therapeutic agents \citep{jiang2021could}. 

The development plan also includes implementing advanced functionalities for identifying compounds that exhibit structural and chemical similarities, as well as pinpointing key fragments crucial for biological activity. This feature will allow researchers to explore a vast chemical space more efficiently, identifying lead compounds with higher precision. These additions are expected to significantly accelerate and deepen the agent's ability to understand compound behaviors in 3D spaces and allow researchers to develop more comprehensive and effective workflows for drug discovery and materials design. Additionally, we plan to include tools that identify key fragments and compounds with similar structural and chemical features from the vast available chemical databases. Tools which can calculate physio-chemical, pharmacokinetic properties, and about sixty other descriptors will be added to the agent to identify quantitative structure-activity relationship (QSAR) and quantitative structure-property relationship (QSPR) to help us with screening the compounds and identifying toxic groups.

Beyond these technical enhancements, there's a focus on making CACTUS more explainable and capable of symbolic reasoning. The aim is to address common criticisms of LLMs, particularly their struggle with reasoning and providing explainable outputs. By integrating more advanced symbolic reasoning capabilities, CACTUS will not only become more powerful in its predictive and analytical functions but also provide users with understandable, logical explanations for its recommendations and predictions. This feature would automate the process of predicting how small molecules, such as drug candidates, interact with targets like proteins, thereby providing invaluable insights into the potential efficacy of new compounds.

The applications of CACTUS extend beyond drug discovery and can be leveraged in other domains such as chemistry, catalysis, and materials science. In the field of catalysis, CACTUS could aid in the discovery and optimization of novel catalysts by predicting their properties and performance based on their structural and chemical features \citep{goldsmith2018machine}. Similarly, in materials science, CACTUS could assist in the design of new materials with desired properties by exploring the vast chemical space and identifying promising candidates for further experimental validation \citep{agrawal2016perspective}.

The future development of CACTUS is geared towards creating an intelligent, comprehensive cheminformatics tool for molecular discovery that not only aids in the identification and design of therapeutic drugs but also ensures a high degree of safety and efficacy. Through the integration of advanced computational techniques and models, alongside improvements in usability and explainability, CACTUS is set to become an indispensable resource in the quest for novel, effective, and safe therapeutic agents, as well as in the discovery and optimization of catalysts and materials.

\section{Conclusions}

In this paper, we have introduced CACTUS, an innovative open-source agent that leverages the power of large language models and cheminformatics tools to revolutionize the field of drug discovery and molecular property prediction. By integrating a wide range of computational tools and models, CACTUS provides a comprehensive and user-friendly platform for researchers and chemists to explore the vast chemical space for molecular discovery and identify promising compounds for therapeutic applications.

We assessed CACTUS performance using various open-source LLMs, including Gemma-7b, Falcon-7b, MPT-7b, Llama2-7b, and Mistral-7b, across a set of one thousand chemistry questions. Our findings indicate that CACTUS outperforms baseline LLMs significantly, with the Gemma-7b and Mistral-7b models achieving the highest accuracy regardless of the prompting strategy employed. Additionally, we investigated the impact of domain-specific prompting and hardware configurations on model performance, highlight the importance of prompt engineering and the potential for deploying smaller models on consumer-grade hardware without significant loss in accuracy. The ability to achieve accurate results with smaller models such Phi on consumer-grade hardware opens up the possibility of wider adoption and accessibility of CACTUS, even for researchers with limited computational resources.

One of the key takeaways from the development and benchmarking of CACTUS is the importance of addressing the challenges associated with model deployment and prompt engineering. The solutions implemented in this work, such as the use of vLLM for hosting models and the development of tailored prompts for each open-source LLM, serve as a valuable foundation for future efforts in this field. As the field of AI continues to evolve rapidly, it is essential to keep abreast of new developments in language modeling and related technologies to further enhance the capabilities and performance of CACTUS. The development and benchmarking of CACTUS also highlighted key challenges in integrating open-source LLMs with domain-specific tools, such as optimizing inference speed and developing effective prompting strategies. We discussed the solutions implemented to address these challenges, including the use of vLLM for model hosting and the creation of tailored prompts for each LLM.

Looking ahead, the future of CACTUS is incredibly promising, with the potential to transform not only drug discovery but also various other domains such as chemistry, catalysis, and materials science. The integration of advanced physics-based AI/ML models, such as 3D-scaffold, reinforcement learning and graph neural networks, will enable a deeper understanding of compound behaviors in 3D spaces, leading to more accurate predictions of molecular interactions and the efficacy and safety of potential therapeutic agents. Moreover, the addition of tools for identifying key fragments, calculating molecular properties, and screening compounds for toxic groups will significantly enhance the efficiency and precision of the drug discovery process. The focus on improving the explainability and symbolic reasoning capabilities of CACTUS will address common criticisms of large language models and provide users with understandable, logical explanations for the tool's recommendations and predictions.

As CACTUS continues to evolve and integrate with other computational tools and autonomous discovery platforms, it has the potential to revolutionize the way we approach drug discovery, catalyst design, and materials science. By leveraging the power of AI and machine learning, CACTUS can help researchers navigate the vast parameter spaces associated with complex chemical systems, identifying promising candidates for experimental validation and optimization. The future development of CACTUS is geared towards creating an intelligent, comprehensive cheminformatics tool that ensures a high degree of safety and efficacy in the identification and design of therapeutic drugs, catalysts, and materials for various application. Through the integration of advanced computational techniques and models, alongside improvements in usability and explainability, CACTUS is set to become an indispensable resource for researchers across various scientific disciplines.

In summary, CACTUS represents a significant milestone in the field of cheminformatics, offering a powerful and adaptable tool for researchers engaged in drug discovery, molecular property prediction, and beyond. As we continue to advance  AI-driven scientific discovery, agent like CACTUS will play a pivotal role in shaping the future of research, innovation, and human health. By embracing the potential of open-source language models and cheminformatics tools, we can accelerate the pace of scientific advancement and unlock new frontiers in the quest for novel, effective, and safe therapeutic agents, catalysts, and materials.

\section*{Code and Data Availability}

The code to run CACTUS and the associated benchmark data can be found on GitHub: \href{https://github.com/pnnl/cactus}{https://github.com/pnnl/cactus}.

\section*{Acknowledgments}

This research was supported by the I3T Investment, under the Laboratory Directed Research and Development (LDRD) Program at Pacific Northwest National Laboratory (PNNL). The computational work was performed using PNNL's research computing at Pacific Northwest National Laboratory. The initial concept of integrating LLM and tools received support from the Exascale Computing Project (17-SC-20-SC), a collaborative effort of two U.S. Department of Energy organizations (Office of Science and the National Nuclear Security Administration) responsible for the planning and preparation of a capable exascale ecosystem, including software, applications, hardware, advanced system engineering, and early testbed platforms, in support of the nation’s exascale computing imperative. PNNL is a multi-program national laboratory operated for the U.S. Department of Energy (DOE) by Battelle Memorial Institute under Contract No. DE-AC05-76RL01830.

%The initial concept of integrating cheminformatics tools received partial support from Federal funding by the NCI-DOE Collaboration established by the U.S. Department of Energy (DOE) and the National Cancer Institute (NCI) of the National Institutes of Health, Cancer Moonshot Task Order No. 75N91019F00134 and under Frederick National Laboratory for Cancer Research Contract 75N91019D00024.

%This work was supported by the U.S. Department of Energy, Office of Science, Advanced Scientific Computing Research, Cancer Deep Learning Environment (CANDLE) project under contract number DE-AC02-06CH11357. 

%This research was supported by the Exascale Computing Project (17-SC-20-SC) Cancer Deep Learning Environment (CANDLE), a collaborative effort of two U.S. Department of Energy organizations (Office of Science and the National Nuclear Security Administration) responsible for the planning and preparation of a capable exascale ecosystem, including software, applications, hardware, advanced system engineering, and early testbed platforms, in support of the nation’s exascale computing imperative.

%This work was supported by the U.S. Department of Energy, Office of Science, Advanced Scientific Computing Research, under contract number 

\section*{Conflict of Interest}

The authors declare that the research was conducted in the absence of any commercial or financial relationships that could be construed as a potential conflict of interest.

\bibliographystyle{unsrtnat}
\setcitestyle{square,numbers,comma}
\bibliography{references}

\begin{thebibliography}{46}
\providecommand{\natexlab}[1]{#1}
\providecommand{\url}[1]{\texttt{#1}}
\expandafter\ifx\csname urlstyle\endcsname\relax
  \providecommand{\doi}[1]{doi: #1}\else
  \providecommand{\doi}{doi: \begingroup \urlstyle{rm}\Url}\fi

\bibitem[Chiesurin et~al.(2023)Chiesurin, Dimakopoulos, Cabezudo, Eshghi,
  Papaioannou, Rieser, and Konstas]{Chiesurin2023}
Sabrina Chiesurin, Dimitris Dimakopoulos, Marco Antonio~Sobrevilla Cabezudo,
  Arash Eshghi, Ioannis Papaioannou, Verena Rieser, and Ioannis Konstas.
\newblock {The Dangers of trusting Stochastic Parrots: Faithfulness and Trust
  in Open-domain Conversational Question Answering}, may 2023.
\newblock URL \url{http://arxiv.org/abs/2305.16519}.

\bibitem[Mialon et~al.(2023)Mialon, Dessì, Lomeli, Nalmpantis, Pasunuru,
  Raileanu, Rozière, Schick, Dwivedi-Yu, Celikyilmaz, Grave, LeCun, and
  Scialom]{mialon2023}
Grégoire Mialon, Roberto Dessì, Maria Lomeli, Christoforos Nalmpantis, Ram
  Pasunuru, Roberta Raileanu, Baptiste Rozière, Timo Schick, Jane Dwivedi-Yu,
  Asli Celikyilmaz, Edouard Grave, Yann LeCun, and Thomas Scialom.
\newblock Augmented language models: a survey, 2023.

\bibitem[Xu et~al.(2023)Xu, Hong, Li, Hu, Chen, and Zhang]{xu2023toolbench}
Qiantong Xu, Fenglu Hong, Bo~Li, Changran Hu, Zhengyu Chen, and Jian Zhang.
\newblock On the tool manipulation capability of open-source large language
  models, 2023.

\bibitem[Qin et~al.(2023)Qin, Liang, Ye, Zhu, Yan, Lu, Lin, Cong, Tang, Qian,
  Zhao, Hong, Tian, Xie, Zhou, Gerstein, Li, Liu, and Sun]{qin2023toolllm}
Yujia Qin, Shihao Liang, Yining Ye, Kunlun Zhu, Lan Yan, Yaxi Lu, Yankai Lin,
  Xin Cong, Xiangru Tang, Bill Qian, Sihan Zhao, Lauren Hong, Runchu Tian,
  Ruobing Xie, Jie Zhou, Mark Gerstein, Dahai Li, Zhiyuan Liu, and Maosong Sun.
\newblock Toolllm: Facilitating large language models to master 16000+
  real-world apis, 2023.

\bibitem[Cai et~al.(2023)Cai, Wang, Ma, Chen, and Zhou]{Cai2023}
Tianle Cai, Xuezhi Wang, Tengyu Ma, Xinyun Chen, and Denny Zhou.
\newblock Large language models as tool makers, 5 2023.

\bibitem[Parisi et~al.(2022)Parisi, Zhao, and Fiedel]{Parisi2022}
Aaron Parisi, Yao Zhao, and Noah Fiedel.
\newblock {TALM: Tool Augmented Language Models}, may 2022.
\newblock URL \url{http://arxiv.org/abs/2205.12255}.

\bibitem[Huang et~al.(2023)Huang, Chen, Mishra, Zheng, Yu, Song, and
  Zhou]{huang2023large}
Jie Huang, Xinyun Chen, Swaroop Mishra, Huaixiu~Steven Zheng, Adams~Wei Yu,
  Xinying Song, and Denny Zhou.
\newblock Large language models cannot self-correct reasoning yet.
\newblock \emph{arXiv preprint arXiv:2310.01798}, 2023.

\bibitem[Kambhampati(2024)]{kambhampati2024can}
Subbarao Kambhampati.
\newblock Can large language models reason and plan?
\newblock \emph{Annals of the New York Academy of Sciences}, 1534\penalty0
  (1):\penalty0 15--18, 2024.

\bibitem[Bommasani et~al.(2021)Bommasani, Hudson, Adeli, Altman, Arora, von
  Arx, Bernstein, Bohg, Bosselut, Brunskill, et~al.]{ontherisk}
Rishi Bommasani, Drew~A Hudson, Ehsan Adeli, Russ Altman, Simran Arora, Sydney
  von Arx, Michael~S Bernstein, Jeannette Bohg, Antoine Bosselut, Emma
  Brunskill, et~al.
\newblock On the opportunities and risks of foundation models.
\newblock \emph{arXiv preprint arXiv:2108.07258}, 2021.

\bibitem[OpenAI(2023)]{gpt4}
OpenAI.
\newblock Gpt-4 technical report, 2023.

\bibitem[Touvron et~al.(2023)Touvron, Lavril, Izacard, Martinet, Lachaux,
  Lacroix, Rozière, Goyal, Hambro, Azhar, Rodriguez, Joulin, Grave, and
  Lample]{llama}
Hugo Touvron, Thibaut Lavril, Gautier Izacard, Xavier Martinet, Marie-Anne
  Lachaux, Timothée Lacroix, Baptiste Rozière, Naman Goyal, Eric Hambro,
  Faisal Azhar, Aurelien Rodriguez, Armand Joulin, Edouard Grave, and Guillaume
  Lample.
\newblock Llama: Open and efficient foundation language models, 2023.

\bibitem[Team et~al.(2024)Team, Mesnard, Hardin, Dadashi, Bhupatiraju, Pathak,
  Sifre, Rivi{\`e}re, Kale, Love, et~al.]{team2024gemma}
Gemma Team, Thomas Mesnard, Cassidy Hardin, Robert Dadashi, Surya Bhupatiraju,
  Shreya Pathak, Laurent Sifre, Morgane Rivi{\`e}re, Mihir~Sanjay Kale,
  Juliette Love, et~al.
\newblock Gemma: Open models based on gemini research and technology.
\newblock \emph{arXiv preprint arXiv:2403.08295}, 2024.

\bibitem[Team(2023)]{MosaicML2023Introducing}
MosaicML~NLP Team.
\newblock Introducing mpt-7b: A new standard for open-source, commercially
  usable llms, 2023.
\newblock URL \url{www.mosaicml.com/blog/mpt-7b}.
\newblock Accessed: 2023-05-05.

\bibitem[Almazrouei et~al.(2023)Almazrouei, Alobeidli, Alshamsi, Cappelli,
  Cojocaru, Debbah, Goffinet, Hesslow, Launay, Malartic,
  et~al.]{almazrouei2023falcon}
Ebtesam Almazrouei, Hamza Alobeidli, Abdulaziz Alshamsi, Alessandro Cappelli,
  Ruxandra Cojocaru, M{\'e}rouane Debbah, {\'E}tienne Goffinet, Daniel Hesslow,
  Julien Launay, Quentin Malartic, et~al.
\newblock The falcon series of open language models.
\newblock \emph{arXiv preprint arXiv:2311.16867}, 2023.

\bibitem[Jiang et~al.(2023)Jiang, Sablayrolles, Mensch, Bamford, Chaplot,
  de~las Casas, Bressand, Lengyel, Lample, Saulnier, et~al.]{jiang2023mistral}
AQ~Jiang, A~Sablayrolles, A~Mensch, C~Bamford, DS~Chaplot, D~de~las Casas,
  F~Bressand, G~Lengyel, G~Lample, L~Saulnier, et~al.
\newblock Mistral 7b (2023).
\newblock \emph{arXiv preprint arXiv:2310.06825}, 2023.

\bibitem[Chiang et~al.(2024)Chiang, Zheng, Sheng, Angelopoulos, Li, Li, Zhang,
  Zhu, Jordan, Gonzalez, et~al.]{chiang2024chatbot}
Wei-Lin Chiang, Lianmin Zheng, Ying Sheng, Anastasios~Nikolas Angelopoulos,
  Tianle Li, Dacheng Li, Hao Zhang, Banghua Zhu, Michael Jordan, Joseph~E
  Gonzalez, et~al.
\newblock Chatbot arena: An open platform for evaluating llms by human
  preference.
\newblock \emph{arXiv preprint arXiv:2403.04132}, 2024.

\bibitem[Zheng et~al.(2023)Zheng, Chiang, Sheng, Zhuang, Wu, Zhuang, Lin, Li,
  Li, and Xing]{zheng2023judging}
L~Zheng, WL~Chiang, Y~Sheng, S~Zhuang, Z~Wu, Y~Zhuang, Z~Lin, Z~Li, D~Li, and
  E~Xing.
\newblock Judging llm-as-a-judge with mt-bench and chatbot arena. arxiv
  preprint arxiv: 230605685.
\newblock 2023.

\bibitem[Hendrycks et~al.(2020)Hendrycks, Burns, Basart, Zou, Mazeika, Song,
  and Steinhardt]{hendrycks2020measuring}
Dan Hendrycks, Collin Burns, Steven Basart, Andy Zou, Mantas Mazeika, Dawn
  Song, and Jacob Steinhardt.
\newblock Measuring massive multitask language understanding.
\newblock \emph{arXiv preprint arXiv:2009.03300}, 2020.

\bibitem[Boiko et~al.(2023)Boiko, MacKnight, and Gomes]{agent1}
Daniil~A. Boiko, Robert MacKnight, and Gabe Gomes.
\newblock Emergent autonomous scientific research capabilities of large
  language models, 2023.

\bibitem[Berthold et~al.(2007)Berthold, Cebron, Dill, Gabriel, K\"{o}tter,
  Meinl, Ohl, Sieb, Thiel, and Wiswedel]{KNIME}
Michael~R. Berthold, Nicolas Cebron, Fabian Dill, Thomas~R. Gabriel, Tobias
  K\"{o}tter, Thorsten Meinl, Peter Ohl, Christoph Sieb, Kilian Thiel, and
  Bernd Wiswedel.
\newblock {KNIME}: The {K}onstanz {I}nformation {M}iner.
\newblock In \emph{Studies in Classification, Data Analysis, and Knowledge
  Organization (GfKL 2007)}. Springer, 2007.
\newblock ISBN 978-3-540-78239-1.

\bibitem[Goecks et~al.(2010)Goecks, Nekrutenko, Taylor, and The~Galaxy]{Galaxy}
Jeremy Goecks, Anton Nekrutenko, James Taylor, and Team The~Galaxy.
\newblock Galaxy: a comprehensive approach for supporting accessible,
  reproducible, and transparent computational research in the life sciences.
\newblock \emph{Genome Biology}, 11\penalty0 (8):\penalty0 R86, 2010.
\newblock ISSN 1474-760X.
\newblock \doi{10.1186/gb-2010-11-8-r86}.
\newblock URL \url{https://doi.org/10.1186/gb-2010-11-8-r86}.

\bibitem[Bran et~al.(2023)Bran, Cox, White, and Schwaller]{chemcrow}
Andres~M Bran, Sam Cox, Andrew~D White, and Philippe Schwaller.
\newblock Chemcrow: Augmenting large-language models with chemistry tools, 4
  2023.

\bibitem[Chase(2022)]{Chase_LangChain_2022}
Harrison Chase.
\newblock {LangChain}, October 2022.
\newblock URL \url{https://github.com/langchain-ai/langchain}.

\bibitem[Wolf et~al.(2020)Wolf, Debut, Sanh, Chaumond, Delangue, Moi, Cistac,
  Rault, Louf, Funtowicz, Davison, Shleifer, von Platen, Ma, Jernite, Plu, Xu,
  {Le Scao}, Gugger, Drame, Lhoest, and Rush]{Wolf_HF}
Thomas Wolf, Lysandre Debut, Victor Sanh, Julien Chaumond, Clement Delangue,
  Anthony Moi, Pierric Cistac, Tim Rault, Remi Louf, Morgan Funtowicz, Joe
  Davison, Sam Shleifer, Patrick von Platen, Clara Ma, Yacine Jernite, Julien
  Plu, Canwen Xu, Teven {Le Scao}, Sylvain Gugger, Mariama Drame, Quentin
  Lhoest, and Alexander Rush.
\newblock {Transformers: State-of-the-Art Natural Language Processing}.
\newblock In \emph{Proceedings of the 2020 Conference on Empirical Methods in
  Natural Language Processing: System Demonstrations}, pages 38--45,
  Stroudsburg, PA, USA, 2020. Association for Computational Linguistics.
\newblock \doi{10.18653/v1/2020.emnlp-demos.6}.
\newblock URL \url{https://www.aclweb.org/anthology/2020.emnlp-demos.6}.

\bibitem[Karpas et~al.(2022)Karpas, Abend, Belinkov, Lenz, Lieber, Ratner,
  Shoham, Bata, Levine, Leyton-Brown, et~al.]{karpas2022mrkl}
Ehud Karpas, Omri Abend, Yonatan Belinkov, Barak Lenz, Opher Lieber, Nir
  Ratner, Yoav Shoham, Hofit Bata, Yoav Levine, Kevin Leyton-Brown, et~al.
\newblock Mrkl systems: A modular, neuro-symbolic architecture that combines
  large language models, external knowledge sources and discrete reasoning.
\newblock \emph{arXiv preprint arXiv:2205.00445}, 2022.

\bibitem[Yao et~al.(2022)Yao, Zhao, Yu, Du, Shafran, Narasimhan, and
  Cao]{Yao2022}
Shunyu Yao, Jeffrey Zhao, Dian Yu, Nan Du, Izhak Shafran, Karthik Narasimhan,
  and Yuan Cao.
\newblock {ReAct: Synergizing Reasoning and Acting in Language Models}, oct
  2022.
\newblock URL \url{http://arxiv.org/abs/2210.03629}.

\bibitem[Landrum et~al.(2013)]{landrum2013rdkit}
Greg Landrum et~al.
\newblock Rdkit: A software suite for cheminformatics, computational chemistry,
  and predictive modeling.
\newblock \emph{Greg Landrum}, 8\penalty0 (31.10):\penalty0 5281, 2013.

\bibitem[Virtanen et~al.(2020)Virtanen, Gommers, Oliphant, Haberland, Reddy,
  Cournapeau, Burovski, Peterson, Weckesser, Bright, et~al.]{virtanen2020scipy}
Pauli Virtanen, Ralf Gommers, Travis~E Oliphant, Matt Haberland, Tyler Reddy,
  David Cournapeau, Evgeni Burovski, Pearu Peterson, Warren Weckesser, Jonathan
  Bright, et~al.
\newblock Scipy 1.0: fundamental algorithms for scientific computing in python.
\newblock \emph{Nature methods}, 17\penalty0 (3):\penalty0 261--272, 2020.

\bibitem[Kim et~al.(2023)Kim, Chen, Cheng, Gindulyte, He, He, Li, Shoemaker,
  Thiessen, Yu, et~al.]{kim2023pubchem}
Sunghwan Kim, Jie Chen, Tiejun Cheng, Asta Gindulyte, Jia He, Siqian He,
  Qingliang Li, Benjamin~A Shoemaker, Paul~A Thiessen, Bo~Yu, et~al.
\newblock Pubchem 2023 update.
\newblock \emph{Nucleic acids research}, 51\penalty0 (D1):\penalty0
  D1373--D1380, 2023.

\bibitem[Davies et~al.(2015)Davies, Nowotka, Papadatos, Dedman, Gaulton,
  Atkinson, Bellis, and Overington]{davies2015chembl}
Mark Davies, Micha{\l} Nowotka, George Papadatos, Nathan Dedman, Anna Gaulton,
  Francis Atkinson, Louisa Bellis, and John~P Overington.
\newblock Chembl web services: streamlining access to drug discovery data and
  utilities.
\newblock \emph{Nucleic acids research}, 43\penalty0 (W1):\penalty0 W612--W620,
  2015.

\bibitem[Irwin et~al.(2020)Irwin, Tang, Young, Dandarchuluun, Wong,
  Khurelbaatar, Moroz, Mayfield, and Sayle]{irwin2020zinc20}
John~J Irwin, Khanh~G Tang, Jennifer Young, Chinzorig Dandarchuluun, Benjamin~R
  Wong, Munkhzul Khurelbaatar, Yurii~S Moroz, John Mayfield, and Roger~A Sayle.
\newblock Zinc20—a free ultralarge-scale chemical database for ligand
  discovery.
\newblock \emph{Journal of chemical information and modeling}, 60\penalty0
  (12):\penalty0 6065--6073, 2020.

\bibitem[Daina and Zoete(2016)]{daina2016boiled}
Antoine Daina and Vincent Zoete.
\newblock A boiled-egg to predict gastrointestinal absorption and brain
  penetration of small molecules.
\newblock \emph{ChemMedChem}, 11\penalty0 (11):\penalty0 1117--1121, 2016.

\bibitem[Li et~al.(2023)Li, Zhao, Yu, Song, Li, Yu, Li, Huang, and Li]{li2023}
Minghao Li, Yingxiu Zhao, Bowen Yu, Feifan Song, Hangyu Li, Haiyang Yu, Zhoujun
  Li, Fei Huang, and Yongbin Li.
\newblock Api-bank: A comprehensive benchmark for tool-augmented llms, 2023.

\bibitem[Farn and Shin(2023)]{farn2023}
Nicholas Farn and Richard Shin.
\newblock Tooltalk: Evaluating tool-usage in a conversational setting, 2023.

\bibitem[Gen(2023)]{Gentopia}
Gentopia.
\newblock \url{https://github.com/Gentopia-AI/Gentopia}, 2023.

\bibitem[Liu et~al.(2023)Liu, Zheng, Du, Ding, Qian, Yang, and
  Tang]{liu2023gpt}
Xiao Liu, Yanan Zheng, Zhengxiao Du, Ming Ding, Yujie Qian, Zhilin Yang, and
  Jie Tang.
\newblock Gpt understands, too.
\newblock \emph{AI Open}, 2023.

\bibitem[Joshi et~al.(2021{\natexlab{a}})Joshi, Gebauer, Bontha, Khazaieli,
  James, Brown, and Kumar]{joshi20213d}
Rajendra~P Joshi, Niklas~WA Gebauer, Mridula Bontha, Mercedeh Khazaieli,
  Rhema~M James, James~B Brown, and Neeraj Kumar.
\newblock 3d-scaffold: A deep learning framework to generate 3d coordinates of
  drug-like molecules with desired scaffolds.
\newblock \emph{The Journal of Physical Chemistry B}, 125\penalty0
  (44):\penalty0 12166--12176, 2021{\natexlab{a}}.

\bibitem[McNaughton et~al.(2022)McNaughton, Bontha, Knutson, Pope, and
  Kumar]{McNaughton2022}
Andrew~D. McNaughton, Mridula~S. Bontha, Carter~R. Knutson, Jenna~A. Pope, and
  Neeraj Kumar.
\newblock {De novo design of protein target specific scaffold-based Inhibitors
  via Reinforcement Learning}.
\newblock may 2022.
\newblock URL \url{http://arxiv.org/abs/2205.10473}.

\bibitem[Knutson et~al.(2022)Knutson, Bontha, Bilbrey, and
  Kumar]{knutson2022decoding}
Carter Knutson, Mridula Bontha, Jenna~A Bilbrey, and Neeraj Kumar.
\newblock Decoding the protein--ligand interactions using parallel graph neural
  networks.
\newblock \emph{Scientific reports}, 12\penalty0 (1):\penalty0 7624, 2022.

\bibitem[Joshi et~al.(2021{\natexlab{b}})Joshi, McNaughton, Thomas, Henry,
  Canon, McCue, and Kumar]{joshi2021quantum}
Rajendra~P Joshi, Andrew McNaughton, Dennis~G Thomas, Christopher~S Henry,
  Shane~R Canon, Lee~Ann McCue, and Neeraj Kumar.
\newblock Quantum mechanical methods predict accurate thermodynamics of
  biochemical reactions.
\newblock \emph{ACS omega}, 6\penalty0 (14):\penalty0 9948--9959,
  2021{\natexlab{b}}.

\bibitem[Joshi et~al.(2023)Joshi, Schultz, Wilson, Kruel, Varikoti, Kombala,
  Kneller, Galanie, Phillips, Zhang, et~al.]{joshi2023ai}
Rajendra~P Joshi, Katherine~J Schultz, Jesse~William Wilson, Agustin Kruel,
  Rohith~Anand Varikoti, Chathuri~J Kombala, Daniel~W Kneller, Stephanie
  Galanie, Gwyndalyn Phillips, Qiu Zhang, et~al.
\newblock Ai-accelerated design of targeted covalent inhibitors for sars-cov-2.
\newblock \emph{Journal of Chemical Information and Modeling}, 63\penalty0
  (5):\penalty0 1438--1453, 2023.

\bibitem[Varikoti et~al.(2023)Varikoti, Schultz, Kombala, Kruel, Brandvold,
  Zhou, and Kumar]{varikoti2023integrated}
Rohith~Anand Varikoti, Katherine~J Schultz, Chathuri~J Kombala, Agustin Kruel,
  Kristoffer~R Brandvold, Mowei Zhou, and Neeraj Kumar.
\newblock Integrated data-driven and experimental approaches to accelerate lead
  optimization targeting sars-cov-2 main protease.
\newblock \emph{Journal of Computer-Aided Molecular Design}, 37\penalty0
  (8):\penalty0 339--355, 2023.

\bibitem[Joshi and Kumar(2021)]{Joshi2021}
Rajendra~P. Joshi and Neeraj Kumar.
\newblock {Artificial Intelligence for Autonomous Molecular Design: A
  Perspective}.
\newblock \emph{Molecules}, 26\penalty0 (22):\penalty0 6761, nov 2021.
\newblock ISSN 1420-3049.
\newblock \doi{10.3390/molecules26226761}.
\newblock URL \url{http://arxiv.org/abs/2102.06045
  https://www.mdpi.com/1420-3049/26/22/6761}.

\bibitem[Jiang et~al.(2021)Jiang, Wu, Hsieh, Chen, Liao, Wang, Shen, Cao, Wu,
  and Hou]{jiang2021could}
Dejun Jiang, Zhenxing Wu, Chang-Yu Hsieh, Guangyong Chen, Ben Liao, Zhe Wang,
  Chao Shen, Dongsheng Cao, Jian Wu, and Tingjun Hou.
\newblock Could graph neural networks learn better molecular representation for
  drug discovery? a comparison study of descriptor-based and graph-based
  models.
\newblock \emph{Journal of cheminformatics}, 13:\penalty0 1--23, 2021.

\bibitem[Goldsmith et~al.(2018)Goldsmith, Esterhuizen, Liu, Bartel, and
  Sutton]{goldsmith2018machine}
Bryan~R. Goldsmith, Jacques Esterhuizen, Jin‐Xun Liu, Christopher~J. Bartel,
  and Christopher Sutton.
\newblock {Machine learning for heterogeneous catalyst design and discovery}.
\newblock \emph{AIChE Journal}, 64\penalty0 (7):\penalty0 2311--2323, jul 2018.
\newblock ISSN 0001-1541.
\newblock \doi{10.1002/aic.16198}.
\newblock URL
  \url{https://aiche.onlinelibrary.wiley.com/doi/10.1002/aic.16198}.

\bibitem[Agrawal and Choudhary(2016)]{agrawal2016perspective}
Ankit Agrawal and Alok Choudhary.
\newblock Perspective: Materials informatics and big data: Realization of the
  “fourth paradigm” of science in materials science.
\newblock \emph{Apl Materials}, 4\penalty0 (5), 2016.

\end{thebibliography}

\appendix
\section{Benchmark Data}
\label{app:bench_data}

\begin{table}[h]
\begin{tabular}{@{}llllll@{}}
\toprule
Model     & Node        & Question Set & Prompt  & Time      & Accuracy \\ \midrule
gemma7b   & A100 80GB   & Quantitative & Full    & 25.61667  & 98.4     \\
gemma7b   & A100 80GB   & Quantitative & Minimal & 36.2      & 96.4     \\
gemma7b   & A100 80GB   & Qualitative  & Minimal & 23.06667  & 95.2     \\
gemma7b   & A100 80GB   & Qualitative  & Full    & 24.93333  & 95.6     \\
gemma7b   & A100 80GB   & Full         & Full    & 43.1      & 94.7     \\
gemma7b   & A100 80GB   & Full         & Minimal & 84.46667  & 85       \\
falcon7b  & A100 80GB   & Full         & Full    & 42.8      & 71.6     \\
mpt7b     & A100 80GB   & Quantitative & Full    & 34.76667  & 71.4     \\
mpt7b     & A100 80GB   & Quantitative & Minimal & 52.25     & 41.6     \\
falcon7b  & A100 80GB   & Full         & Minimal & 72.91667  & 16.8     \\
llama27b  & A100 80GB   & Full         & Full    & 187.46667 & 64.3     \\
llama27b  & A100 80GB   & Full         & Minimal & 31.66667  & 66.3     \\
mistral7b & A100 80GB   & Full         & Minimal & 44.51667  & 88.9     \\
mistral7b & A100 80GB   & Full         & Full    & 39.45     & 97       \\
phi2      & V100        & Quantitative & Minimal & 19.93333  & 93.6     \\
phi2      & RTX 2080 Ti & Quantitative & Minimal & 22.18333  & 93       \\
phi2      & A100 80GB   & Quantitative & Minimal & 12.1833   & 93.6     \\
phi2      & A100 80GB   & Full         & Minimal & 30.21667  & 69.3     \\
phi2      & A100 80GB   & Full         & Full    & 49.05     & 85.4     \\
olmo1b    & RTX 2080 Ti & Full         & Full    & 49.48333  & 52.2     \\
olmo1b    & RTX 2080 Ti & Full         & Minimal & 51.48333  & 0        \\ \bottomrule
\end{tabular}
\end{table}

\end{document}